\documentclass[sigconf]{acmart}

\usepackage{amsmath}
\usepackage{amssymb}
\usepackage{multirow}
\usepackage{hyperref}
\usepackage[inline]{enumitem}
\usepackage{subfigure}
\usepackage{bm}
\usepackage{amsmath}
\usepackage{booktabs}
\usepackage{graphicx}
\usepackage{balance}

\AtBeginDocument{%
  \providecommand\BibTeX{{%
    \normalfont B\kern-0.5em{\scshape i\kern-0.25em b}\kern-0.8em\TeX}}}

\copyrightyear{2021}
\acmYear{2021}
\setcopyright{acmlicensed}\acmConference[SIGIR '21]{Proceedings of the 44th International ACM SIGIR Conference on Research and Development in Information Retrieval}{July 11--15, 2021}{Virtual Event, Canada}
\acmBooktitle{Proceedings of the 44th International ACM SIGIR Conference on Research and Development in Information Retrieval (SIGIR '21), July 11--15, 2021, Virtual Event, Canada}
\acmPrice{15.00}
\acmDOI{10.1145/3404835.3463074}
\acmISBN{978-1-4503-8037-9/21/07}

\acmSubmissionID{123-A56-BU3}

\begin{document}

\title{Graph Pooling via Coarsened Graph Infomax
%%Graph Representation Learning with Pooled Graph Infomax Pooling
}

\fancyhead{}
\author{Yunsheng Pang$^{1}$, Yunxiang Zhao$^{2,1}$, Dongsheng Li$^{2,*}$}
 \affiliation{
   \institution{$^{1}$The University of Melbourne}
}
 \affiliation{
   \institution{$^{2}$National Lab for Parallel and Distributed Processing, College of Computer, National University of Defense Technology}
}
\email{{yunshengp, yunxiangz}@student.unimelb.edu.au, lds1201@163.com}
\thanks{*Dongsheng Li is the corresponding author.}
\renewcommand{\shortauthors}{Yunsheng Pang, Yunxiang Zhao, Dongsheng Li}

\begin{abstract}
	Graph pooling that summaries the information in a large graph into a compact form is essential in hierarchical graph representation learning. Existing graph pooling methods either suffer from high computational complexity or cannot capture the global dependencies between graphs before and after pooling. To address the problems of existing graph pooling methods, we propose \textbf{C}oarsened \textbf{G}raph \textbf{I}nfomax \textbf{Pool}ing (\textbf{CGIPool}) that maximizes the mutual information between the input and the coarsened graph of each pooling layer to preserve graph-level dependencies. To achieve mutual information neural maximization, we apply contrastive learning and propose a self-attention-based algorithm for learning positive and negative samples. Extensive experimental results on seven datasets illustrate the superiority of CGIPool comparing to the state-of-the-art methods.
\end{abstract}


% <concept_id>10010147.10010257.10010293.10010319</concept_id>
% <concept_desc>Computing methodologies~Learning latent representations</concept_desc>

% \ccsdesc[500]{Computing methodologies~Learning latent representations}

\maketitle
\section{Introduction}
%proved to be more efficient when processing graph structured data, such as social networks, biological networks and chemical molecules ~\cite{lazer2009life,davidson2002genomic,duvenaud2015convolutional}, and
Graph Neural Networks (GNNs) ~\cite{kipf2016semi,hamilton2017inductive,velivckovic2017graph} have shown outstanding performance in numerous graph related tasks, such as node classification~\cite{kipf2016semi}, graph classification~\cite{zhang2018end} and link prediction~\cite{liben2007link}. Recently, learning representations of entire graphs~\cite{zhang2018end,ying2018hierarchical} has attracted a lot of attention in the fields of bioinformatics~\cite{feragen2013scalable}, recommendation systems~\cite{fu2020fairness,su2021detecting}, social network study~\cite{zhuang2017understanding}, etc~\cite{lu2020vgcn}. In graph level representation learning, graph pooling~\cite{ying2018hierarchical} that maps large graphs to smaller ones is essential for capturing a meaningful structure of the graph and reducing the computation cost simultaneously. 

\par Earlier graph pooling methods use summation or neural networks to pool all the representations of nodes~\cite{atwood2015diffusion,gilmer2017neural,zhang2018end}. However, these methods cannot well capture graph structure information such as subgraph and hierarchy. In recent years, \textit{hierarchical pooling} methods have been proposed to address the limitations of global pooling. In hierarchical pooling, \textit{graph coarsening pooling} methods consider graph pooling as a node clustering problem, which merges similar nodes to a cluster and regards the cluster centers as the nodes in the coarsened graph ~\cite{ying2018hierarchical,yuan2020structpool,bianchi2019mincut}. These methods perform node clustering by learning a soft cluster assignment matrix~\cite{ying2018hierarchical}, which cannot leverage the sparsity in the graph topology. Moreover, the high computational complexity of these methods prevents them from being applied to large graphs. \textit{Node selection pooling} methods have lower complexity compared with graph coarsening ~\cite{gao2019graph,lee2019self,ranjan2020asap,li2020graph}. These methods select a subset of nodes according to the importance of nodes in the graph. However, they neglect the global dependencies between graphs before and after pooling and consider nodes' local topology~\cite{lee2019self} only when learning nodes' importance scores. Therefore, they cannot filter out nodes that are informative locally but not essential for graph-level representation learning.
\par % To tackle the problems above, we propose a node-selection-based graph pooling model named CGIPool, 
To tackle the problems of node selection pooling, we propose CGIPool that maximizes the mutual information~\cite{belghazi2018mine,hjelm2018learning,velickovic2019deep} between the input and the coarsened graph of each pooling layer. Specifically, we first propose positive and negative coarsening modules with the self-attention mechanism to learn real and fake coarsened graphs. The real coarsened graph, which maximally reflects the input graph, is used as the positive sample. The fake coarsened graph, which contains unimportant nodes of the input graph, is used as the negative sample. Then, we train a discriminator~\cite{creswell2018generative,zhao2019cbhe} to increase the score on the positive example and decrease the score on the negative example so as to achieve mutual information neural maximization. Compared with other node selection pooling models, CGIPool that considers the global dependencies between graphs before and after pooling can preserve nodes that are informative globally rather than those informative locally. We summarize the main contributions of this paper as follows:

%\par CGIPool inherits the advantages of node-selection-based pooling methods, hence has a lower complexity than graph-coarsening-based methods. 
\begin{itemize}[leftmargin=*]
	\item We propose CGIPool, which is the first node selection pooling model that captures global dependencies between the input and the coarsened graph of each pooling layer.
	\item We propose self-attention-based positive and negative coarsening modules to learn real and fake coarsened graphs as positive and negative samples, thus be able to achieve mutual information neural maximization via contrastive learning.
	\item Experimental results demonstrate that CGIPool outperforms state-of-the-art methods on six out of seven benchmark datasets. We have made the implementation of CGIPool public available\footnote{The source code is public available at: https://github.com/PangYunsheng8/CGIPool}.
\end{itemize}

\section{Related Work}
We review existing works on graph pooling, followed by those on mutual information neural estimation and maximization.
\par \textbf{Graph Pooling:} Graph pooling can be divided into global pooling and hierarchical pooling ~\cite{ranjan2020asap}. Global pooling models summarize a graph into a node directly~\cite{atwood2015diffusion,gilmer2017neural,zhang2018end}. Hierarchical pooling that captures the local substructures of graphs can be divided into graph-coarsening-based, node-selection-based methods, and a few others. (i) Graph-coarsening pooling aims to map nodes into different clusters~\cite{ying2018hierarchical,yuan2020structpool,bianchi2019mincut}. (ii) Node-selection pooling adopts learnable scoring functions to drop nodes with lower scores~\cite{gao2019graph,lee2019self,ranjan2020asap,li2020graph}.
Apart from (i) and (ii), \cite{diehl2019edge} scores edges and merges nodes with high-scoring edges between them, and ~\cite{khasahmadi2020memory} proposes a memory layer to jointly learn node representations and coarsen the input graph.

%Graph pooling can be divided into global pooling and hierarchical pooling. In global pooling, Set2Set~\cite{gilmer2017neural} performs global pooling by aggregating information through LSTMs~\cite{hochreiter1997long}. DGCNN summarizes
%the graph by concatenating a subset of nodes after sorting them in descending order. Hierarchical graph pooling can further be divided into graph-coarsening-based and node-selection-based methods. In graph-coarsening-based pooling,
%considers graph pooling as a node clustering problem, which merges similar nodes into clusters and regards the cluster centers as the nodes in the coarsened graph.
%DiffPool~\cite{ying2018hierarchical} proposes to assign nodes to a set of clusters by learning a soft cluster assignment matrix with GNNs, StructPool~\cite{yuan2020structpool} integrates conditional random field when learning the soft cluster assignment matrix. In node-selection-based pooling, TopKPool~\cite{gao2019graph} and SAGPool~\cite{lee2019self} learn the information scores for nodes in a graph using neural networks, ASAP~\cite{ranjan2020asap} first generates clusters by aggregating neighboring nodes and then drops the lower score clusters using a scoring function.
\par \textbf{Mutual information neural estimation:} Mutual information (MI) measures the mutual dependencies between two random variables~\cite{belghazi2018mine}. Deep informax~\cite{hjelm2018learning} simultaneously estimates and maximizes the mutual information between the pixels and overall images to learn image representations. DGI~\cite{velickovic2019deep} maximizes the mutual information between the input graph and each of its nodes to learn informative node representations. VIPool~\cite{li2020graph} leverages MI to select nodes that maximally represent their neighborhoods, which is the only graph pooling model based on mutual information maximization. Our CGIPool considers both the information from nodes' local neighborhoods and the global dependencies between the input and the coarsened graph, which is different from VIPool that considers nodes' local neighborhoods only.

\section{Method}
We present the overall structure of the proposed CGIPool in Figure~\ref{CGIPool}, where an input graph is first fed into the proposed positive and negative coarsening modules to learn the real and fake coarsened graphs. Then, the two coarsened graphs along with the input graph are encoded as the positive and negative samples. To maximize the mutual information between the input and the coarsened graph, we first train a discriminator that increases the score on the positive sample and decreases the score on the negative sample~\cite{creswell2018generative,zhao2019cbhe}, we then fuse the real and fake coarsened graphs to obtain the final coarsened graph. Next, we detail the mutual information estimation and maximization followed by the positive and negative coarsening in CGIPool.

\subsection{Mutual Information Maximization} 
\begin{figure*}[h] 
	\vspace{-0.5cm}
	\centering 
	\includegraphics[width=0.9\textwidth]{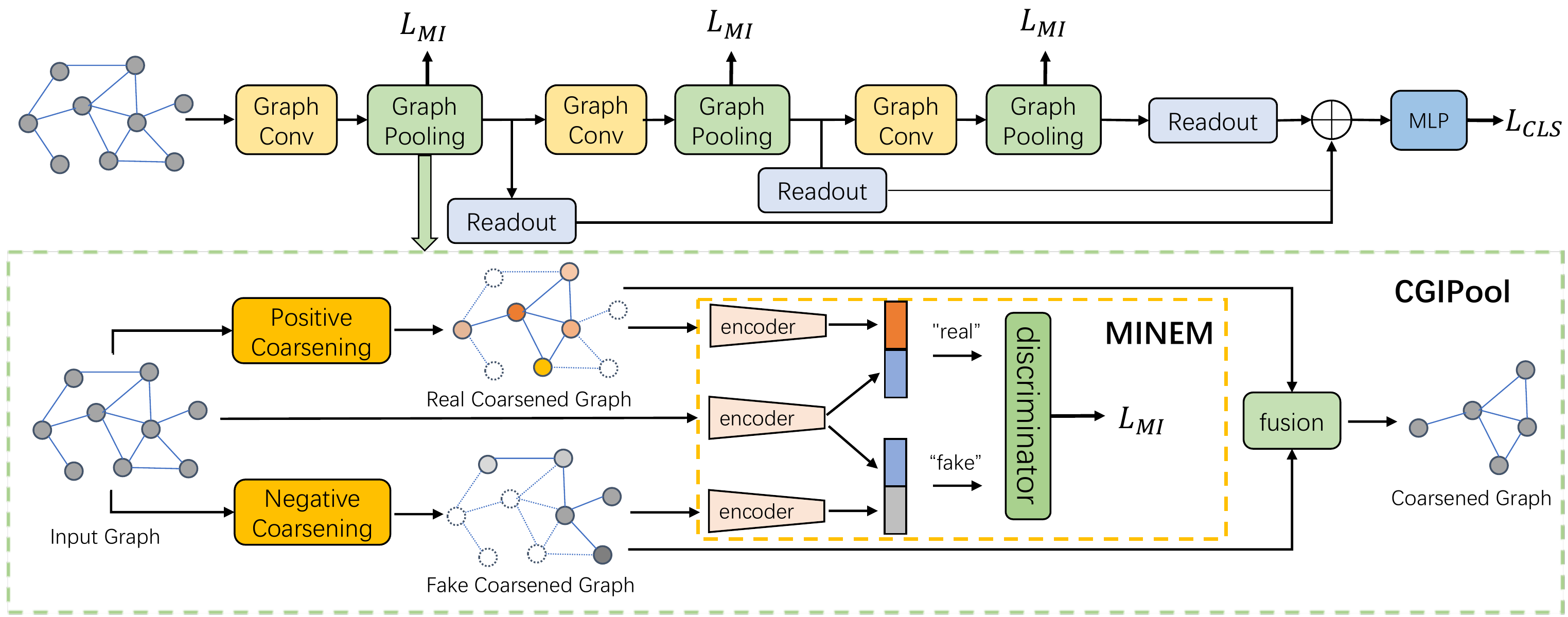} 
	\vspace{-0.3cm}
	\caption{The overall network for graph classification and the proposed CGIPool.}
	\label{CGIPool}  
	\vspace{-0.4cm}
\end{figure*}

Given a graph $G(\mathcal{V}, \mathcal{E}, X)$ with $N=|\mathcal{V}|$  vertices and $|\mathcal{E}|$ edges, $X \in \mathbb{R}^{N \times d}$ denotes the node feature matrix, $A \in \mathbb{R}^{N \times N}$ denotes the adjacency matrix. We denote $G^l$ and $G^{l+1}$ be the input and the coarsened graph of $l$-th pooling layer in the model. Let random variable $\mathbf{O}$ be the representation of the input graph of each pooling layer, e.g., $G^l$. Another variable $\mathbf{Z}$ be the representation of the coarsened graph, e.g., $G^{l+1}$. The distribution of $\mathbf{O}$ and $\mathbf{Z}$ are $P_O=P(O=e^l=encoder(H^l))$ and $P_Z=P(Z=e^{l+1}=encoder(H^{l+1}))$, where $H^l \in \mathbb{R}^{N \times f}$ is the feature matrix of $G^l$ and $H^0=X$. $f$ is the dimension of node feature matrix, $e^l \in \mathbb{R}^{1 \times f}$ is the representation of $G^l$, $encoder$ is a function that maps $G^l$ to a vector, e.g., a $readout$ function. The mutual information between the input graph $G^l$ and the coarsened graph $G^{l+1}$ is computed as the KL-divergence between the joint distribution $P_{\mathbf{O}, \mathbf{Z}}$ of $P_{\mathbf{O}}$ and $P_{\mathbf{Z}}$, and the product of marginal distribution $P_{\mathbf{O}} \otimes P_{\mathbf{Z}}$ ~\cite{belghazi2018mine}:
\begin{equation}
\setlength{\abovedisplayskip}{-0.00pt}
\setlength{\belowdisplayskip}{-0.00pt}
\begin{aligned}
    I(O, Z) &=D_{KL}(P_{O, Z}||P_{O} \otimes P_{Z}) \\
    & \geq \mathop{sup}\limits_{T \in \mathcal{T}} \{\mathbb{E}_{e^l,e_{r}^{l+1} \sim P_{O,Z}}[T(e^l,e_{r}^{l+1})] \\
    &- \mathbb{E}_{e^l \sim P_{O},e_{f}^{l+1} \sim P_{Z}}[e^{T(e^l,e_{f}^{l+1})-1}] \}
\end{aligned}
\label{eq1}
\end{equation}
where $I(O,Z)$ is the mutual information between $O$ and $Z$, and its lower bound is learned via contrastive learning ~\cite{saunshi2019theoretical}. Function $T \in \mathcal{T}$ maps a pair of input graph and coarsened graph to a real value, reflecting the dependency of the two graphs. $e_r^{l+1}$ and $e_f^{l+1}$ are the coarsened graphs that sampled from $P_{\mathbf{O}, \mathbf{Z}}$ and $P_{\mathbf{O}} \otimes P_{\mathbf{Z}}$, respectively. In practice, we replace the KL divergence by a GAN-like divergence to achieve mutual information maximization~\cite{dziugaite2015training}:
%where $I(O,Z)$ is the mutual information between $O$ and $Z$, $T \in \mathcal{T}$ is an arbitrary function that maps a pair of input graph and coarsened graph to a real value, reflecting the dependency of the two graphs. $e_r^{l+1}$ and $e_f^{l+1}$ are the coarsened graphs that sampled from $P_{\mathbf{O}, \mathbf{Z}}$ and $P_{\mathbf{O}} \otimes P_{\mathbf{Z}}$, respectively. In practice, the KL divergence can be replaced by a GAN-like divergence so as to achieve mutual information maximization via contrastive learning:%training a discriminator:
\begin{equation}
\setlength{\abovedisplayskip}{-0.00pt}
\setlength{\belowdisplayskip}{-0.00pt}
\begin{aligned}
I_{GAN}{(O,Z)} & \geq \mathop{sup}\limits_{T \in \mathcal{T}} \{\mathbb{E}_{ P_{O,Z}}[\log\sigma(T_{\theta}(e^l,e_{r}^{l+1}))] \\
    &+ \mathbb{E}_{P_{O},P_{Z}}[\log(1-\sigma(T_{\theta}(e^l,e_{f}^{l+1})))] \}
\end{aligned}
\end{equation}
where $\sigma$ is the sigmoid function. We parameterize $T$ in Equation~\ref{eq1} with a discriminator network $T_{\theta}$. Then the mutual information between the input and the coarsened graph of each pooling layer can be maximized by minimizing the mutual information loss $\mathcal{L}_{MI}$:
\begin{equation}
\setlength{\abovedisplayskip}{-0.00pt}
\setlength{\belowdisplayskip}{-0.00pt}
    \mathcal{L}_{MI}=-\frac{1}{N}\frac{1}{L_p}\sum_{i=1}^{N}\sum_{l=1}^{L_p} [{\log \sigma(T_{\theta}{(e_i^l,e_{r,i}^{l+1})})\!+\! \log(1\!-\!\sigma(T_{\theta}{(e_i^l,e_{f,i}^{l+1})}))}]
\end{equation} 
where $L_p$ is the number of pooling layers, $N$ is the size of the training set. The yellow square in Figure~\ref{CGIPool} shows the structure of mutual information neural estimation and maximization (MINEM). 
\par In Figure~\ref{CGIPool}, the overall model for graph classification repeats the graph convolution and pooling operations three times. The overall graph level representation is the sum of coarsened graphs from different pooling layers, which will be fed into an MLP layer with a softmax classifier. The overall loss is:
\begin{equation}
\setlength{\abovedisplayskip}{-0.00pt}
\setlength{\belowdisplayskip}{-0.00pt}
    \mathcal{L} = \mathcal{L}_{CLS} + \alpha \mathcal{L}_{MI}
    \label{final_loss}
\end{equation}
where $\alpha$ is a hyperparameter that balances the two terms. $\mathcal{L}_{CLS}$ is the graph classification loss:
\begin{equation}
\setlength{\abovedisplayskip}{-0.00pt}
\setlength{\belowdisplayskip}{-0.00pt}
    \mathcal{L}_{CLS}=-\sum_{i=1}^{N}\sum_{j=1}^{C} {y_{ij}\log\hat{y}_{ij}}
\end{equation}
where $y_{ij}$ is the ground-truth label and $\hat{y_{ij}}$ is the predicted probability that the graph belongs to class $i$, $C$ is the number of classes.

\subsection{Positive and Negative Coarsening Module}
%In this section, we present the detail of generating $e_r^{l+1}$ and $e_f^{l+1}$, and introduce a fusion operation to obtain the coarsened graph $G^{l+1}$, which is the output of $l$-th pooling layer.
We detail the positive and negative coarsening modules to learn real and fake coarsened graphs based on self-attention mechanism. Specifically, the positive coarsening module selects a subset of nodes in $G^l$ to form a real coarsened graph $G_{r}^{l+1}$ which maximally represent $G^l$, while the negative coarsening module generates a fake coarsened graph $G_{f}^{l+1}$ which contains unimportant nodes in $G^l$. Then $G_{r}^{l+1}$ and $G_{f}^{l+1}$ are fed into an encoder network to obtain their representations $e_r^{l+1}$ and $e_f^{l+1}$. We also present the fusion operation that obtains the coarsened graph $G^{l+1}$ from $G_{r}^{l+1}$ and $G_{f}^{l+1}$, which is the output of $l$-th pooling layer.
\par In the positive and negative coarsening modules, we first adopt two parallel graph neural networks to calculate a 1D attention score vector for each node in the input graph $G^l$:
\begin{equation}
\setlength{\abovedisplayskip}{-0.00pt}
\setlength{\belowdisplayskip}{-0.00pt}
\begin{split}
y_r^{l+1} &= \sigma(GNN_r^{l}(H^l, A^l))\\
y_f^{l+1} &= \sigma(GNN_f^{l}(H^l, A^l))
\end{split}
\end{equation}
where $\sigma$ is the sigmoid function. $H^l$ and $A^l$ are the node feature matrix and adjacency matrix of $G^l$. $y_r^{l+1} \in \mathbb{R}^{N \times 1}$ and $y_f^{l+1} \in \mathbb{R}^{N \times 1}$ are score vectors learned by $\text{GNN}_r^{l}$ and $\text{GNN}_f^{l}$. 
We then rank all nodes according to the score vectors $y_r^{l+1}$ and $y_f^{l+1}$ and select the $k$-largest values:
\begin{equation}
\setlength{\abovedisplayskip}{-0.00pt}
\setlength{\belowdisplayskip}{-0.00pt}
\label{eq:idx}
\begin{split}
idx_r^{l+1} &= \text{top-k}(y_r^{l+1}, k) \\
idx_f^{l+1} &= \text{top-k}(y_f^{l+1}, k)
\end{split}
\end{equation}
where $k$ is the number of selected nodes, $idx_r^{l+1}$ contains the indices of nodes selected for the real coarsened graph $G_{r}^{l+1}$, and $idx_f^{l+1}$ contains the indices of nodes selected for the fake coarsened graph $G_{f}^{l+1}$. Based on $idx_r^{l+1}$ and $idx_f^{l+1}$, the adjacency matrices $A_r^{l+1}$ of $G_{r}^{l+1}$ and $A_f^{l+1}$ of $G_{f}^{l+1}$ are:
\begin{equation}
\setlength{\abovedisplayskip}{-0.00pt}
\setlength{\belowdisplayskip}{-0.00pt}
\begin{split}
A_{r}^{l+1} &= A^l(idx_r^{l+1}, idx_r^{l+1}) \\
A_{f}^{l+1} &= A^l(idx_f^{l+1}, idx_f^{l+1})
\end{split}
\end{equation}
and the feature matrices $H_r^{l+1}$ of $G_{r}^{l+1}$ and $H_f^{l+1}$ of $G_{f}^{l+1}$ are:
\begin{equation}
\setlength{\abovedisplayskip}{-0.00pt}
\setlength{\belowdisplayskip}{-0.00pt}
\begin{split}
H_{r}^{l+1} &= H^l(idx_r^{l+1}, :) \odot y_r^{l+1}(idx_r^{l+1}, :)\\
H_{f}^{l+1} &= H^l(idx_f^{l+1}, :) \odot y_f^{l+1}(idx_f^{l+1}, :)
\end{split}
\end{equation}
where $\odot$ is the broadcasted elementwise product. Then the representations of the input and two coarsened graphs are learned via weight-shared \emph{encoder} network:
\begin{equation}
\setlength{\abovedisplayskip}{-0.00pt}
\setlength{\belowdisplayskip}{-0.00pt}
\begin{split}
e^l &= \text{encoder}(H^l) \\
e_r^{l+1} &= {encoder}(H_r^{l+1}) \\
e_f^{l+1} &= {encoder}(H_f^{l+1})
\end{split}
\end{equation}
%where the \emph{encoder} networks are weight-shared. 

\par We have presented the procedure to obtain $e_r^{l+1}$ and $e_f^{l+1}$ above. To force $GNN_r^{l+1}$ and $GNN_f^{l+1}$ learn meaningful $y_r^{l+1}$ and $y_f^{l+1}$, where the real coarsened graph $G_r^{l+1}$ can represent $G^{l}$ and the fake coarsened graph $G_f^{l+1}$ contains unimportant nodes in $G^{l}$, we propose a fusion operation:
%Then, we introduce a fusion operation that computes the difference of $y_r^{l+1}$ and $y_f^{l+1}$:
\begin{equation}
\setlength{\abovedisplayskip}{-0.00pt}
\setlength{\belowdisplayskip}{-0.00pt}
y_{d}^{l+1} = \sigma(y_r^{l+1} - y_f^{l+1})
\end{equation}
where $\sigma$ is the sigmoid function that normalizes the score values. Based on the self-attention mechanism, important nodes tend to have larger values in $y_r^{l+1}$ and smaller values in $y_f^{l+1}$.
We then select top-k nodes $idx_{d}^{l+1} \&= \text{top-k}(y_{d}^{l+1}, k)$
%according to $y_{d}^{l+1}$ 
to form the coarsened graph $G^{l+1}$, and the node feature matrix $H^{l+1}$ of $G^{l+1}$ are given by:
\begin{equation}
\setlength{\abovedisplayskip}{-0.00pt}
\setlength{\belowdisplayskip}{-0.00pt}
\begin{split}
%idx_{d}^{l+1} &= \text{top-k}(y_{d}^{l+1}, k) \\
H^{l+1} &= H^l(idx_{d}^{l+1}, :) \odot y_{d}^{l+1}(idx_{d}^{l+1}, :)
\end{split}
\end{equation}
%$idx_{d}^{l+1}$ contains the indices of nodes selected for the coarsened graph, 
Therefore, the real coarsened graph $G_r^{l+1}$ formed by the nodes with the $k$-largest values in $y_r^{l+1}$ can represent the input graph $G^l$, and the fake coarsened graph $G_f^{l+1}$ formed by the nodes with the $k$-largest values in $y_f^{l+1}$ contains unimportant nodes in $G^l$.

%with larger values in $y_r^{l+1}$ and smaller values in $y_f^{l+1}$ are important nodes and will be selected to form $G^{l+1}$. %Therefore, important nodes will be assigned to larger values by $\text{GNN}_r^{l}$ and unimportant nodes will be assigned to larger values by $\text{GNN}_f^{l}$, respectively.
%\par We then rank all nodes in the graph according to score vectors $y_r^{l+1}$ and $y_f^{l+1}$ and select the $k$-largest values:

\begin{table*}[htb]
	\vspace{-0.3cm}
	\setlength{\abovecaptionskip}{0.cm}
	\setlength{\belowcaptionskip}{-0.cm}
	\begin{center}
		\centering
		\caption{The statistics of datasets and graph classification accuracies of different graph pooling algorithms.}
		\setlength{\tabcolsep}{2.8mm}{
			\begin{tabular}{cccccccc}
				\hline
				\multirow{2}{*}{Dataset}& \multicolumn{3}{c}{Small Molecules}&
				\multicolumn{3}{c}{Social Networks}& \multicolumn{1}{c}{Bioinformatics}\\
				
				\cmidrule(r){2-4} \cmidrule(r){5-7} \cmidrule(r){8-8}  
				&NCI1& NCI109& Mutagenicity& IMDB-B& IMDB-M& COLLAB& PROTEINS\\
				\hline
				\# Graphs (Classes)& 4110 (2)& 4127 (2)& 4337 (2)& 1000(2)& 1500(3)& 5000(3)& 1113(2)\\
				Avg \# Nodes& 29.87& 29.68& 30.32& 19.77& 13.00& 74.49& 39.06\\
				Avg \# Edges& 32.30& 32.13& 30.77& 96.53& 65.94& 2457.78& 72.82\\
				\hline
				Set2Set~\cite{gilmer2017neural}& 74.82$\pm$0.85& 74.12$\pm$1.31& 77.69$\pm$0.55& 70.98$\pm$0.78& 48.64$\pm$0.46& 77.62$\pm$0.38& 71.46$\pm$2.17\\
				DGCNN~\cite{zhang2018end}& 74.98$\pm$0.96& 74.43$\pm$1.26& 77.95$\pm$0.67& 70.82$\pm$0.66& 48.71$\pm$0.58& 77.85$\pm$0.51& 70.89$\pm$2.23\\
				DiffPool~\cite{ying2018hierarchical}& 77.12$\pm$0.98& 76.43$\pm$1.42& 80.44$\pm$0.82& 73.20$\pm$0.85& 50.57$\pm$0.83& 78.78$\pm$0.72& 71.26$\pm$2.66\\
				TopKPool~\cite{gao2019graph}& 76.22$\pm$1.14& 75.26$\pm$1.35& 79.14$\pm$0.76& 71.68$\pm$0.87& 49.32$\pm$0.72& 77.98$\pm$0.63& 72.61$\pm$2.23\\
				SAGPool~\cite{lee2019self}& 76.84$\pm$1.21& 75.98$\pm$1.47& 79.18$\pm$0.82& 72.20$\pm$0.91& 49.92$\pm$0.69& 78.56$\pm$0.77& 73.16$\pm$2.31\\
				ASAP~\cite{ranjan2020asap}& 77.21$\pm$1.35& 76.33$\pm$1.58& 80.12$\pm$0.88& 72.74$\pm$0.93& 50.28$\pm$0.61& 78.95$\pm$0.69& 73.86$\pm$2.52\\
				VIPool~\cite{li2020graph}& 77.49$\pm$1.81& 76.73$\pm$1.62& 80.19$\pm$1.02& {73.37$\pm$1.41}& 50.49$\pm$0.96& 78.87$\pm$1.23& 73.65$\pm$2.66\\
				\hline
				CGIPool& \textbf{78.62$\pm$1.04}& \textbf{77.94$\pm$1.37}& \textbf{80.65$\pm$0.79}& 72.40$\pm$0.87& \textbf{51.45$\pm$0.65}& \textbf{80.30$\pm$0.69}& \textbf{74.10$\pm$2.31}\\
				\hline
		\end{tabular}}
		\label{table3}
	\end{center}
	\vspace{-0.4cm}
\end{table*}

\section{Experiments}
We compare CGIPool with state-of-the-art models on graph datasets for graph classification tasks. CGIPool is implemented with PyTorch and PyTorch Geometric on an NVIDIA P100 GPU.

\subsection{Datasets and Experimental Setup}
\textbf{Datasets}:
We conduct extensive experiments to evaluate our proposed CGIPool on seven datasets~\cite{morris2020tudataset}, including three small molecules datasets: {NCI1}, {NCI109}~\cite{shervashidze2011weisfeiler} and {Mutagenicity}~\cite{kazius2005derivation}, three social network datasets: {IMDB-B}, {IMDB-M} and {COLLAB}~\cite{yanardag2015structural}, and a protein graph dataset: {PROTEINS}~\cite{feragen2013scalable}. Table~\ref{table3} summarizes the statistics of these datasets. Since no node feature is provided in social network datasets, we use one-hot vectors to encode the degrees of nodes as features, which is the same as~\cite{li2020graph}.

\textbf{Baselines}:
We validate the effectiveness of the proposed method by comparing it with state-of-the-art graph pooling models, including 
two global pooling methods: Set2Set~\cite{gilmer2017neural} and DGCNN~\cite{zhang2018end};  DiffPool~\cite{ying2018hierarchical} which is the first differentiable graph pooling method; and
four node-selection-based graph pooling methods: TopKPool~\cite{gao2019graph}, SAGPool~\cite{lee2019self}, ASAP~\cite{ranjan2020asap}, and VIPool~\cite{li2020graph}.

\textbf{Experimental setup}: We randomly split each dataset into three parts: 80\% for training, 10\% for validation, and 10\% for testing~\cite{lee2019self}. We repeat this random splitting 20 times and take the averaged performance with standard derivation as the final results. We set the dimension of node embedding as 128, the learning rate as 0.001, the weight decays as 0.0001, and the pooling ratio $r$ as 0.8 for all datasets.
%\{0.1, 0.01, 0.001, $1e^{-4}$, $1e^{-5}$\}. 
We adopt Adam as the optimizer and stop the training when the validation loss does not decrease for 100 consecutive epochs.  The discriminator consists of a two-layer MLP followed by a sigmoid function. The hyper-parameter $\alpha$ in Eq(\ref{final_loss}) is set to 1 for the PROTEINS dataset, and 0.001 for all rest datasets. 
All baselines and CGIPool are equipped with the same three-layer GCN model as shown in Figure~\ref{CGIPool}, and pooling modules are the only differences.
%We also note that to make the comparison fairer, all results shown in Table \ref{table3} are obtained by keeping all structures (e.g., the network architecture and hyperparameters) unchanged and only change the pooling modules.

\subsection{Results}
Table~\ref{table3} summarizes the accuracy of CGIPool and baselines on graph classification. CGIPool achieves state-of-the-art performance on all datasets except the IMDB-B dataset, where the accuracy of CGIPool is only 0.97\% lower than that of the best baseline. Moreover, the datasets cover different domains such as social networks, bioinformatics, and molecules, which reveals that CGIPool yields great performance in different domains. Specifically, CGIPool outperforms state-of-the-art models by up to 1.21\%, 1.43\%, and 0.45\% on social networks, bioinformatics, and molecules datasets, respectively.

\subsection{Ablation and Robustness study}
We examine the impact of the negative coarsening module and the mutual information maximization in CGIPool. We also examine the robustness of CGIPool with different pooling ratios.

\subsubsection{Negative coarsening module}
The negative coarsening module in CGIPool aims to learn the fake coarsened graphs. To illustrate its impact, we implement CGIPool-RS, which is a variant of CGIPool that replacing the negative coarsening module by \textbf{R}andomly \textbf{S}elect the same number of nodes from the input graph. The selected nodes form a random fake coarsened graph and will be fed into the discriminator. Table~\ref{table4} shows that CGIPool outperforms CGIPool-RS consistently, which validates that the proposed negative coarsening module can learn fake coarsened graphs that better reflect the distribution of $P_{\mathbf{O}} \otimes P_{\mathbf{Z}}$.

\begin{table}[tbh]
\setlength{\abovecaptionskip}{0.cm}
\setlength{\belowcaptionskip}{-0.cm}
	\begin{center}
		\centering
		\caption{Results of CGIPool and its two variants.}
		\setlength{\tabcolsep}{1pt}{
		\begin{tabular}{ccccc}
			\hline
			Dataset& NCI1& NCI109& Mutagenicity& COLLAB\\
			\hline
			CGIPool& \textbf{78.62$\pm$1.04}& \textbf{77.94$\pm$1.37}& \textbf{80.65$\pm$0.79}& \textbf{80.30$\pm$0.69}\\
			CGIPool-RS& 77.21$\pm$1.35& 76.33$\pm$1.58& 80.12$\pm$0.88& 79.14$\pm$0.82\\
			CGIPool w/o MI& 76.92$\pm$1.30& 75.65$\pm$1.44& 79.23$\pm$0.91& 78.3$\pm$0.75\\
			\hline
		\end{tabular}}
		\label{table4}
	\end{center}
    \vspace{-0.4cm}
\end{table}

%we conduct two baselines in this section. The first is SAGPool, which does not contain a negative pooling module and mutual information maximization. The second is RS-NP, where the fake pooled graph is randomly selected from the graph before pooling with the same number of nodes as XXX. 
\subsubsection{Mutual Information Estimation and Maximization}
To demonstrate the impact of maximizing the mutual information between the input and the coarsened graph of each pooling layer, we remove the mutual information maximization in CGIPool, and denote this variant as "CGIPool w/o MI". In Table~\ref{table4}, CGIPool outperforms CGIPool w/o MI consistently, which proves the importance of maximizing the mutual information between the graph before and after each pooling layer.
%, which are used for generating negative samples of mutual information neural estimation.
%we design experiments to verify the validity of the proposed negative pooling module. In the experiments, 
%we replace the negative pooling module with a random approach, where random select the same number of nodes from the graph before pooling as the random selection fake pooled graph, then the pooled graphs are used as negative samples to train the discriminator. Table \ref{table4} shows the results of random selection based negative pooling (RS-NP) and our self-attention based negative pooling (SA-NP) on for datasets. We also add a baseline that does not have any negative pooling modules and mutual information maximization.
%\par Table \ref{table4} demonstrates the superiority of the proposed method from two aspects. First, the results of random-selection based negative pooling are better than the baseline, which proves to maximize the mutual information between the graph before and after each pooling layer can preserve more useful information, therefore, the proposed method outperforms other methods that consider local neighborhoods only. Second, replacing random-selection based negative pooling module with the proposed negative pooling module obtains better results, which validates the negative pooling module can learn a representation that better reflects the true distribution of $P_{\mathbf{O}} \otimes P_{\mathbf{Z}}$.

\subsubsection{Performance with different pooling ratio} 
\begin{figure}[tbh]
\vspace{-0.3cm}
\setlength{\abovecaptionskip}{0.cm}
\setlength{\belowcaptionskip}{-0.2cm}
	\centering
	\subfigure[NCI1]{
		\begin{minipage}{0.47\linewidth}
			\centering
			\includegraphics[width=1.6in]{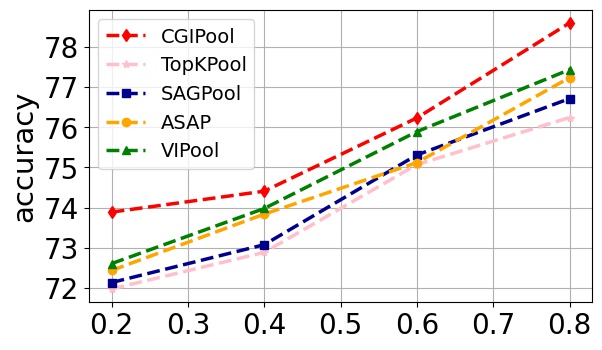}
		\end{minipage}
		\label{NCI1-pr}
	}\,\,
	\subfigure[IMDB-M]{
		\begin{minipage}{0.47\linewidth}
			\centering
			\includegraphics[width=1.6in]{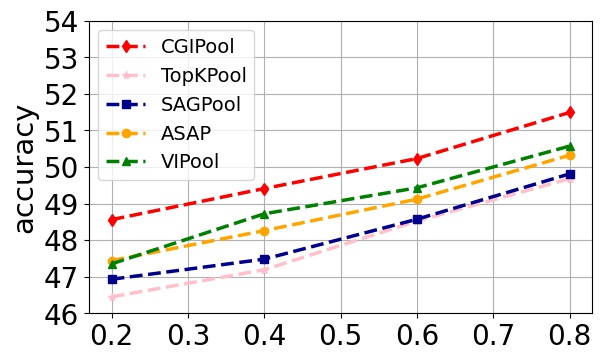}
		\end{minipage}
		\label{IMDBM-pr}
	}
    \vspace{-0.1cm}
	\caption{The performance of four pooling models on NCI1 and IMDB-M with different pooling ratios.} 
	\label{vis}
    \vspace{-0.4cm}
\end{figure}

To prove the robustness of CGIPool under different pooling ratios, we summarize the accuracy of CGIPool and three node selection based pooling methods on NCI1 and IMDB-M datasets in Figure~\ref{vis}, where the pooling ratio varies from 0.2 to 0.8 with a step length of 0.2.
%CGIPool considers the global dependencies between the graph before and after each pooling layer and preserves more useful information with the same number of nodes. 
CGIPool outperforms all baselines consistently, which proves that CGIPool can learn the nodes that are essential for graph-level representation learning regardless of the pooling ratio. 
%Figure~\ref{vis} shows the performance of four pooling models on NCI1 and IMDB-M with different pooling ratios. The performance of the proposed method outperforms others in all different pooling ratios, which proves the proposed method can find the nodes that are essential for graph-level representation learning accurately, even if the number of selected nodes is small. 

\section{Conclusion}  
In this paper, we propose CGIPool that maximizes the mutual information between the input and the coarsened graph of each pooling layer. To achieve contrastive learning for mutual information neural maximization, we propose self-attention-based positive and negative coarsening modules to learn real and fake coarsened graphs as positive and negative samples. Extensive experiments show that the proposed CGIPool outperforms state-of-the-art methods on six out of seven graph classification datasets.

\section{Acknowledgments}
This work is sponsored in part by the National Key Research \& Development Program of China under Grant No. 2018YFB0204300, and the National Natural Science Foundation of China under Grant No. 62025208 and 61932001.

\bibliographystyle{IEEEtran}
\balance 
\bibliography{ref.bib} 

\end{document}